\documentclass[10pt, a4paper]{article}

\usepackage{lrec-coling2024}

\usepackage{natbib}
\usepackage{multibib}
\makeatletter
\def\@mb@citenamelist{cite,citep,citet,citealp,citealt,citepalias,citetalias}
\makeatother
\newcites{languageresource}{~}

\usepackage{graphicx}
\usepackage{tabularx}
\usepackage{soul}

\usepackage{booktabs}
\usepackage{subcaption}
\usepackage{multirow}
\usepackage{dcolumn}
\usepackage{siunitx}

\usepackage{enumitem}
\usepackage{verbatim} 
\usepackage{algorithm}
\PassOptionsToPackage{noend}{algpseudocode}
\usepackage{algpseudocode}
\usepackage{amsmath,amssymb}
\newcommand{\todo}[1]{\textcolor{red}{TODO: #1}}
\newcommand{\edit}[1]{\textcolor{blue}{#1}}

\newcommand{\robert}[1]{\textcolor{teal}{#1}}

\usepackage{csquotes}
\usepackage{tikz}
\usepackage{pgfplots}
\pgfplotsset{width=8cm,compat=1.9}
\usepackage{pgf-pie}

\usepackage{xcolor}

\usepackage{hyperref}

 \definecolor{darkblue}{rgb}{0, 0, 0.5}
  \hypersetup{breaklinks=true, colorlinks=true, citecolor=darkblue, linkcolor=darkblue, urlcolor=darkblue}

\usepackage{xstring}

\usepackage{color}

\title{Towards Realistic Few-Shot Relation Extraction:\\ 
A New Meta Dataset and Evaluation}

\name{Fahmida Alam\textsuperscript{1}, Md Asiful Islam\textsuperscript{1}, Robert Vacareanu\textsuperscript{1,2}, Mihai Surdeanu\textsuperscript{1}} 

\address{\textsuperscript{1}University of Arizona, Tucson, USA\\
         \textsuperscript{2}Technical University of Cluj-Napoca, Cluj-Napoca, Romania\\
         \{fahmidaalam, asifulislam, rvacareanu, msurdeanu\}@arizona.edu\\}

\abstract{
We introduce a meta dataset for few-shot relation extraction, which includes two datasets derived from existing supervised relation extraction datasets -- NYT29~\cite{takanobu2019hierarchical,nayak2020effective} and WIKIDATA~\cite{sorokin2017context} -- as well as a few-shot form of the TACRED dataset~\cite{sabo2021revisiting}. Importantly, all these few-shot datasets were generated under realistic assumptions such as: the test relations are different from any relations a model might have seen before, limited training data, and a preponderance of candidate relation mentions that do not correspond to any of the relations of interest. Using this large resource, we conduct a comprehensive evaluation of six recent few-shot relation extraction methods, and observe that no method comes out as a clear winner. Further, the overall performance on this task is low, indicating substantial need for future research.
We release all versions of the data, i.e., both supervised and few-shot, for future research.$^1$ 
 \\ \newline  }

\begin{document}

\maketitleabstract
\footnotetext[1]{Datasets and additional resources are available at: \href{https://github.com/clulab/releases/tree/master/lrec2024-realistic-fewshot-meta-dataset}{Link}}

\section{Introduction}


Information Extraction (IE) plays a pivotal role in Natural Language Processing (NLP). IE is fundamental to many NLP tasks such as question answering, event extraction, knowledge base population, etc. Relation Extraction (RE) is a sub-task of IE with the focus of identifying entities and their semantic relations in a given text, enabling the extraction of structured information from unstructured data. For instance, in the sentence ``John Doe was born in New York City'', Relation Extraction can transform this into a structured tuple such as $\rightarrow$ $(\text{John Doe}, \texttt{born in}, \text{New York City})$, indicating the inherent relation between the person, action, and location.

Many supervised methods have been proposed to address the relation extraction task \cite[inter alia]{soares2019matching, zhang2018graph, wang2016relation, miwa2016end}. However, a traditional supervised machine learning (ML) setup is not always realistic for RE due to the large amount of training data required.
This setup is mostly incompatible with real-world RE scenarios such as pandemic response or intelligence, in which RE models must be developed and deployed quickly with minimal supervision. 

Considering this task setup, a realistic choice for solving this problem is few-shot learning (FSL) and its RE equivalent, few-shot relation extraction (FSRE), in which (a) each relation class is associated with a very small number of examples (typically 1 or 5), and the relation classes in the testing partition are different from any relations a model might have seen before. 
While several FSRE datasets and methods have been proposed recently (see Related Work), this subfield of NLP is still poorly understood due to a lack of realistic datasets and rigorous evaluations. 
This observation has motivated this work, in which we introduce a meta dataset for the task as well as a meaningful evaluation of multiple FSRE methods on this data.
The key contributions of our work are:

\begin{enumerate}[label=(\alph*)]
    \item We develop a meta dataset for FSRE, which includes three datasets: one based on NYT29~\cite{takanobu2019hierarchical,nayak2020effective}, one based on WIKIDATA~\cite{sorokin2017context}, and lastly the few-shot variant of TACRED proposed by~\cite{sabo2021revisiting}.
    All these datasets were converted into realistic few-shot variants using the procedure detailed in \S~\ref{sec:supervised-to-fewshot}. This procedure guarantees a setup that is aligned with real-world applications, e.g., the test relations are different from any relations available in a background dataset, limited training data, preponderance of candidate relation mentions that do not correspond to any of the relations of interest, etc. 
    \item We conduct a comprehensive evaluation of six recent FSRE methods using this meta dataset. Our evaluation reveals that none of the models emerged as a definitive winner. Furthermore, the overall performance of the best models was notably low, indicating the substantial need for future research. Our datasets will contribute as an invaluable resource for this future research.
\end{enumerate}

\section{Related Work}
\subsection{Methods}
Historically,  relation extraction approaches can be categorized as either rule-based or relying on statistical models. In the past decade, the latter category has been dominated by neural-based methods. More recently, hybrid directions have emerged, aiming to combine the advantages of both. We delve deeper into each of these directions.

\subsubsection{Rule-based Methods}
Prior to the widespread adoption of statistical machine learning, rule-based approaches enjoyed a period of prominence. 
These methods typically involve the acquisition of rules representative of specific relations. For example, the rule \texttt{[ne=PER]+ <nsubj born >nmod\_in [ne=LOC]+} captures a syntactic pattern for the \texttt{born\_in} relation, where the pattern matches if the underlying constraints are satisfied: a named entity labeled as person is connected to the word \texttt{born} with a \texttt{nominal subject} dependency, and the same word \texttt{born} is further connected to a named entity labeled as location with a \texttt{nominal modifier} dependency. For example, this pattern will match the sentence: \textit{John Doe was born in New York City.}
A match of such rules is then interpreted as an indication that the two entities participate in the corresponding relation.

In \cite{hearst-1992-automatic}, the authors propose a set of handwritten rules to extract words satisfying the hyponymy relation. Subsequently, efforts were directed toward automating the learning of such patterns \cite{Riloff1993AutomaticallyCA, Riloff1996AutomaticallyGE, Riloff1999LearningDF} with and without supervision. \cite{Gupta2014ImprovedPL} improves automatic pattern learning by allowing soft matching in the form of predicting the labels on unlabeled entities. 

Another prominent line of work for rule-based methods is that of casting the pattern learning problem as a graph-based problem and leveraging graph-based algorithms \cite{Kozareva2008SemanticCL, Vacareanu2022PatternRankJR}.



\subsubsection{Neural-based Methods}
The adoption of neural-based methods has grown significantly due to their high performance, making them the de facto approach for relation extraction tasks today. Many underlying architectures were proposed for relation extraction, such as ones based on CNNs \cite{Zeng2014RelationCV, Nguyen2015RelationEP}, RNNs \cite{Zhang2015RelationCV}, LSTMs \cite{zhang2017position}, or, more recently, Transformers \cite{Vaswani2017AttentionIA, Joshi2019SpanBERTIP}. These approaches typically operate end-to-end and are built on top of pre-trained embeddings, either static \cite{Mikolov2013DistributedRO, Pennington2014GloVeGV, Bojanowski2016EnrichingWV} or contextual \cite{McCann2017LearnedIT, Peters2018DeepCW, Devlin2019BERTPO}.

A more recent direction has been translating the relation extraction task into a different NLP task to leverage more training data \cite{Chen2022NewFO}. For example, relation extraction can be cast as an entailment problem \cite{Sainz2021LabelVA, Rahimi2023ImprovingZR}, or as summarization \cite{Lu2022SummarizationAI}.







A distinctive direction emerged in the last years, attempting to combine the advantages of both rule-based systems and neural-based systems. For example, \citet{vacareanu-etal-2022-examples} teaches a neural network to generate rules for RE. Other directions aiming to improve the explainability of the resulting model include: (i) learning an explainability classifier jointly with the RE model to ensure faithfulness of explanations \cite{Tang2021InterpretabilityRJ, Tang2023BootstrappingNR}, or (ii) learning a neural ``soft'' (or semantic) matcher to improve the rules' recall \cite{Zhou2020NEROAN}.



\subsection{Datasets and Methods for Few-Shot Relation Extraction}
A key contribution to the RE space is the creation of datasets that support the development of new RE approaches.
A recent survey \cite{bassignana-plank-2022-mean} categorized popular relation classification datasets based on their data sources into three main categories: (i) News and Web, (ii) Scientific Publications, and (iii) Wikipedia, totaling 17 datasets. We refer the reader to this survey for more details. 

An important and realistic setting for this task is few-shot relation extraction (FSRE), where only a small number of training examples are available for each relation class to be learned. 
Notably, only three datasets are available in a few-shot format \cite{bassignana-plank-2022-mean}: FewRel \citelanguageresource{Han2018FewRelAL}, FewRel 2.0 \citelanguageresource{Gao2019FewRel2T}, and few-shot TACRED \citelanguageresource{sabo2021revisiting}. 

The FewRel dataset, containing 70,000 sentences covering 100 relations from Wikipedia, is created by identifying relation mentions through distant supervision; noise is subsequently filtered by crowd-workers \cite{Han2018FewRelAL}. Later on, the FewRel 2.0 dataset \cite{Gao2019FewRel2T}, an extension of the original FewRel dataset \cite{Han2018FewRelAL}, introduced a new test set in a distinct domain and included the option of a NOTA (None of the Above) relation.

\citet{sabo2021revisiting} argues that FewRel provides an unrealistic benchmark due to its uniform relation distribution and the prevalence of proper nouns as entities. Although FewRel 2.0 tried to amend it using an updated episode sampling procedure, the evaluation setup is still notably unrealistic \cite{sabo2021revisiting}. 
As a solution, \citet{sabo2021revisiting} converted the supervised TACRED dataset \cite{zhang2017position} into a few-shot TACRED variant by applying realistic episode sampling. Concretely, the episode in an FSRE evaluation should be selected in a way that follows all the criteria (a--f) we mention in Section~\ref{sec:supervised-to-fewshot}. 
To develop our other few-shot datasets, i.e., NYT29 and WIKIDATA, we followed a similar strategy (see \S~\ref{sec:supervised-to-fewshot} and \ref{sec:episodes}). 

Nevertheless, despite the unquestionable contribution of such datasets to the RE field, we observed a lack of consistency in the results observed in the various proposed evaluations. For example, some methods evaluated on FewRel attained an accuracy of 93.9\%, surpassing human-level performance at 92.2\% {\cite{soares2019matching}}. While FewRel 2.0 yields lower results, i.e., the best method achieved 80.3\% \cite{Gao2019FewRel2T}, they are still remarkably high, given the challenging nature of the task.

Further, \cite{sabo2021revisiting} evaluated their MNAV model (which was state-of-art at the time) on FewRel 2.0 and achieved an F1 score of approximately 78\% for 5-way 1-shot and 80\% for 5-way 5-shot, whereas the best results on TACRED are much lower: the F1 score is 12.4\% for 5-way 1-shot and 30.0\% for 5-way 5-shot. These differences are caused by differences in how the datasets are constructed, 
which impacts consistent analyses of the proposed methods. To remedy this issue, we propose a meta dataset for few-shot RE that includes three datasets that are constructed using the same realistic procedure and capture multiple important phenomena. This allows us to rigorously evaluate multiple approaches for few-shot RE as shown in \S~\ref{sec:models}.

\section{Dataset Construction Process}
We detail next our first key contribution: the construction of a meta dataset for FSRE, which combines two new FSRE datasets and a third existing one.

\subsection{Data Sources}
\label{sec:dataSource}
We leverage three existing {\em supervised} datasets for RE to serve as our starting point. These datasets cover a diverse set of domains: NYT29, WIKIDATA, and TACRED.

\paragraph{NYT29:} 
The NYT29 dataset originates from the New York Times corpus, which comprises a collection of more than 1.8 million articles authored and released by the New York Times between January 1, 1987, and June 19, 2007, with article metadata provided by the New York Times Newsroom \citelanguageresource{sandhaus2008}. 
This dataset was annotated with relations from Freebase using distant supervision by \citet{riedel2010modeling}. 
Depending on how many relation classes are kept, this original dataset has multiple versions, e.g., ``NYT10,'' ``NYT11,'' and ``NYT29''  \cite{takanobu2019hierarchical,nayak2020effective}. Our work relies on the latter version, which contains 29 distinct relations (e.g.,{\tt /people/person/place\_lived}), and it covers a wide range of topics, news events, and perspectives.

\paragraph{WIKIDATA:} 
The WIKIDATA dataset is a subset of Wikipedia, wherein articles have been marked with Wikidata relations using distant supervision \citelanguageresource{sorokin2017context}. This corpus encompasses two primary types of annotations: entities and relations. Entity annotations are derived from Wikipedia article links. Each link has been converted to a Wikidata identifier using the mappings from the Wikidata itself. Additional entities are recognized using a named entity recognizer and are linked to Wikidata.

\paragraph{TACRED:} 
Unlike the previous two datasets, which were annotated using distant supervision, TACRED was {\em manually} annotated for 42 relation classes from the TAC KBP challenge \citelanguageresource{SurdeanuHeng:14} (e.g., {\tt per:schools\_attended} and {\tt org: members}) plus {\tt no\_relation}. The dataset contains 106,264 RE examples, which were annotated over textual data from both newswire sources and the corpus employed in the annual TAC Knowledge Base Population (TAC KBP) challenges \citelanguageresource{zhang2017position}. 
These examples are generated by merging human annotations obtained from the TAC KBP challenges and crowdsourcing. 

It is important to note that these datasets capture distinct phenomena that are important for RE: 
{\flushleft {\bf (1)}} NYT29 and WIKIDATA were annotated using distant supervision, whereas TACRED was manually annotated. It is known that distant supervision introduces label noise~\cite{riedel2010modeling}. This is particularly important for the negative class, i.e., in the case of distant supervision, negative labels can be false negatives. That is, they should not be interpreted as ``no known relation label applies'' but rather as ``we have no information about this entity pair in the knowledge base.'' This impacts the sampling procedure discussed later in this section.
{\flushleft {\bf (2)}} NYT29 allows multiple relations to exist between the same two entities in the same sentence. For example, in the sentence ``Mr. Mashal, speaking in Damascus, Syria, said \dots'' and the entity pair ``Damascus'' and ``Syria'' is annotated with two relations: {\tt administrative\_divisions} and {\tt capital}. Because of this, multi-label RE classifiers may have an advantage on NYT29.
{\flushleft{\bf (3)}} WIKIDATA allows for overlapping entities. For example, in the sentence 
``\dots featuring Lon Chaney and Andrew Lloyd Webber's 1986 musical .'' and the entity pair ``1986'' and ``1986 musical'' is annotated with {\tt first performance}. This is likely to confuse methods that rely on entity markers \cite{zhou-chen-2022-improved}.
\subsection{Linguistic Annotations}
Since some of these datasets were not accompanied by linguistic annotations, we processed the texts in house to guarantee that the same linguistic information is available for all three datasets. For all linguistic annotations, we used the {\tt processors} library.\footnote[2]{\url{https://github.com/clulab/processors}} This library uses LSTM-CRFs \cite{lample2016neural} for case restoration, part-of-speech (POS) tagging, named entity recognition (NER), and the method of \citet{vacareanu2020parsing} for dependency parsing. 

\subsubsection{NYT29}
In the original NYT29 dataset, the texts in the three partitions (train, dev, test) were initially presented in lowercase, which led to certain inaccuracies during linguistic annotation. To solve this problem, we first restored case using the LSTM-CRF in the {\tt processors} library. On a small sample, we observed that this restoration is over 95\% accurate.

We then tokenized the text and applied POS tagging, NER, and dependency parsing. However, to determine the subject and object type for each relation mentioned, we used the provided gold entity labels in the original dataset (see Table~\ref{tab:nyt_gold}).

\begin{table}

    \begin{tabular}{p{0.95\columnwidth}}
        \hline
        \textbf{\textit{Sentence:}} ``An arts center that the town of old Saybrook plans to open next year will be named after Katharine Hepburn.''\\\\
        \textit{Entity 1: } ``Katharine Hepburn''\\
        \textit{Predicted label:} PERSON\\
        \textit{Gold label: } PERSON\\\\
        \textit{Entity 2: } ``old Saybrook''\\
        \textit{Predicted label:} ORGANIZATION\\
        \textit{Gold label: } LOCATION\\
        \hline
    \end{tabular}
    \caption{An example from NYT29 with gold and predicted entity labels. We used the gold entity labels for this dataset.}
    \label{tab:nyt_gold}
\end{table}
We observed that a small number of sentences in the NYT29 dataset were not parsed into a dependency tree by the {\tt processors} parser (i.e., the parser produced several subtrees that covered different sentence fragments). The main cause of this error was long and complex sentences. However, the number of sentences with such errors was small: 0.1\% of the training sentences, 0.07\% in dev, and 0.1\% in the test. For simplicity, we removed these sentences from train and dev, and, in order to not modify the test partition, we manually corrected the parse trees for the sentences in the test.

\subsubsection{WIKIDATA}

For WIKIDATA, we used the same NLP library for tokenization, POS tagging, NER, and dependency parsing. Case restoration was not needed for the WIKIDATA sentences. 

However, one important difference between NYT29 and WIKIDATA is that the labels for entities participating in relations in WIKIDATA are limited to just two: ``Lexical'' for named entities, and ``Date'' for dates. To increase the informativeness of entity labels, we adopted the labels predicted by the {\tt processors} NER if they overlap with the span of the entity labels in WIKIDATA. If no predicted NE overlaps with a relation entity, we keep the default WIKIDATA entity label.


\subsubsection{TACRED}
In the TACRED dataset, essential NLP tasks, i.e., POS tagging, NER, and dependency parsing, were performed using Stanford CoreNLP \cite{manning2014stanford} and included in the original dataset. 
To maintain compatibility with previous works, we keep the same linguistic annotations.

\begin{table}[t]
    \centering
    \begin{tabular}{c|c}
        \hline
        Dataset & Entity Labeling Scheme\\
        \hline
        NYT29 & Gold labels\\
        WIKIDATA & Predicted labels \\
        TACRED & Predicted labels \\
        \hline
    \end{tabular}
    \caption{Labeling scheme for entities participating in relations in the three datasets considered.}
    \label{tab:entity_type_labelling}
\end{table}

Importantly, TACRED and our version of WIKIDATA use labels predicted by a NER for the entities participating in a relation, whereas NYT29 uses gold labels. Table~\ref{tab:entity_type_labelling} summarizes this information.

\subsection{Negative Class Label Standarization}
The concept of negative relations refers to instances where the relation between two entities either does not fit into any predefined categories, or it may indicate that there is no relation between them at all. Note that negative labels are handled differently in the three datasets considered:
{\flushleft {\bf (1)}} NYT29 contains no annotations for the negative relation label. In this situation, we introduce negative examples using the supervised-to-few-shot transformation described in Section~\ref{sec:supervised-to-fewshot} and Algorithm 1.
{\flushleft {\bf (2)}} In contrast, TACRED and WIKIDATA explicitly annotate some negative relations between entity pairs that co-occur in the same sentence (TACRED uses the {\tt no\_relation} label, while WIKIDATA uses {\tt P0}). 

The above differences impact the few-shot version of these datasets (see \S~\ref{sec:supervised-to-fewshot}) and, thus, the performance of few-shot RE models. 
Lastly, we standardize the label for negative relations to {\tt NOTA}  across the three datasets. 


\vspace{2mm}
To increase reproducibility, after all these pre-processing steps were applied, we formatted all three datasets using the same tabular format. The format is described in Appendix A. This is the same format the TACRED dataset used. We followed the exact format so that we could apply the transformation technique of converting the supervised dataset into the few-shot dataset described in \cite{sabo2021revisiting}.

\subsection{Supervised to Few-Shot Transformation}
\label{sec:supervised-to-fewshot}

We transform the supervised NYT29, TACRED, and WIKIDATA datasets into FSRE datasets by applying a generalized form of the transformation method described in \cite{sabo2021revisiting}. This process transforms a supervised dataset into a {\em realistic}  FSRE dataset by following a series of constraints that are likely to occur in real-world applications:

{\flushleft {\bf (a)}} The test (or ``target'') relation classes are {\em different} from any of relations that might be available in a background dataset (``background relations'');

{\flushleft {\bf (b)}} The number of training examples $K$ for each target relation class is very small (typically 1 or 5);

{\flushleft {\bf (c)}} The distribution of relations is not uniform, i.e., some relations are rarer than others;

{\flushleft {\bf (d)}} Most candidate relation mentions do not correspond to a target relation;

{\flushleft {\bf (e)}} Many relation candidates seen in testing may not correspond also to a background relation. Thus, a traditional supervised RE classifier that trains on the background data is not applicable;

{\flushleft {\bf (f)}} Entities participating in relations may include named entities, as well as pronouns and common nouns.

Before we formalize the transformation process, we introduce some necessary notations:

\begin{description}
\item[\(\boldsymbol{C}\)] -- A set of known relation classes in a dataset partition.

\item[{\tt NOTA}] -- The relation class {\tt NOTA} (None-of-the-above) is assigned to entity pairs whose corresponding relation class is not in the applicable \(\boldsymbol{C}\) set. Note that this is different from the {\tt no\_relation} label used in the supervised datasets. In the FSRE setting, {\tt NOTA} includes both {\tt no\_relation} examples as well as all positive relation labels that are not used in the dataset partition at hand \cite{sabo2021revisiting}. 

\item[\(\boldsymbol{D}\)] -- A relation classification dataset such that \(D: \{(x_i,c_i)\}_{i=1}^n\), where \(\forall c_i \in C \cup \{\){\tt NOTA}\(\} \).

\item[\(x_i\)] -- Represents the \textit{i-th} instance in a RE dataset \(D\) such that \(x_i = (e_1, e_2, s)_i\) where \(e_1\) and \(e_2\) represent a pair of entities in a sentence \(s\), where the relation between this two entity is labeled \(c_i\).

\item[$N$-Way $K$-Shot] --
We follow the $N$-way $K$-shot setup for FSRE, as proposed by \cite{vinyals2016matching, snell2017prototypical}. In an $N$-way $K$-shot setup, a classifier aims to discriminate between \(N\) target relation classes using only a support set \(K\) examples of each. Typically, \(K\) is a very small number, e.g., 1 or 5.

\end{description}

\begin{algorithm}[t]
\caption{Transformation of a supervised RE dataset to few-shot RE using the $N$-way $K$-shot setup}\label{alg:transformation}
\begin{algorithmic}
\State \textbf{Input: }\(\boldsymbol{D}, \boldsymbol{C}\)
\State \textbf{Output: }\(\boldsymbol{D}_{FS}, \boldsymbol{C}_{FS}\)\\
\State Step 0: Replace {\tt no\_relation} with {\tt NOTA} in $\boldsymbol{D}$; remove {\tt no\_relation} from $\boldsymbol{C}$, if present\\

\State Step 1: Split \(\boldsymbol{D}\) in \(\boldsymbol{D}_{train}\), \(\boldsymbol{D}_{dev}\), and \(\boldsymbol{D}_{test}\)\\

\State Step 2: Choose a random split of \(\boldsymbol{C}\) as \(\boldsymbol{C}_{train}\), \(\boldsymbol{C}_{dev}\), and \(\boldsymbol{C}_{test}\) such that the following two conditions are true: 
\begin{enumerate}[label=(\alph*)]
    \item \(\boldsymbol{C}_{train}\) , \(\boldsymbol{C}_{dev}\), and \(\boldsymbol{C}_{test}\) be pairwise disjoint
    \item  \(|\boldsymbol{C}_{train}|\) , \(|\boldsymbol{C}_{dev}|\), and \(|\boldsymbol{C}_{test}| \geq N\) (for $N$-way $K$-shot)
\end{enumerate}
 \\\\

Step 3:
\For{each \((x_i,c_i) \in \boldsymbol{D}_{train}\)}
    \If{\(c_i \notin \boldsymbol{C}_{train}\)}
        \State $c_i = $ {\tt NOTA}
    \Else
        \State Retain the original label
    \EndIf
\EndFor        
~\\
\State Step 4: Repeat Step 3 for \(\boldsymbol{D}_{dev}\) and \(\boldsymbol{D}_{test}\) using their corresponding \(\boldsymbol{C}_{dev}\), and \(\boldsymbol{C}_{test}\) label sets\\

\State Step 5: 
\(\boldsymbol{C}_{train} = \boldsymbol{C}_{train} \cup \{\){\tt NOTA}\(\}\)\\
~~~~~~~~~~~~~\(\boldsymbol{C}_{dev} = \boldsymbol{C}_{dev} \cup \{\){\tt NOTA}\(\}\)\\
~~~~~~~~~~~~~\(\boldsymbol{C}_{test} = \boldsymbol{C}_{test} \cup \{\){\tt NOTA}\(\}\)\\

\State Step 6:
\(\boldsymbol{C}_{FS} = (\boldsymbol{C}_{train}, \boldsymbol{C}_{dev}, \boldsymbol{C}_{test})\)\\
~~~~~~~~~~~~~\(\boldsymbol{D}_{FS} = (\boldsymbol{D}_{train}, \boldsymbol{D}_{dev}, \boldsymbol{D}_{test})\)\\
\Return \(\boldsymbol{C}_{FS}, \boldsymbol{D}_{FS}\)
\end{algorithmic}
\end{algorithm}

Algorithm \ref{alg:transformation} describes the transformation process of a supervised RE dataset  $\boldsymbol{D}$ containing relation labels $\boldsymbol{C}$ into a few-shot dataset {$\boldsymbol{D}_{FS}$, $\boldsymbol{C}_{FS}$}.
%
The two key steps of the transformation algorithm are as follows. First, we split the original dataset into three partitions (train/dev/test) such that they are pairwise disjoint with respect to the positive relations they contain (steps 1 and 2). For example, if the train partition contains the relation {\tt country of origin}, this relation is not allowed to appear in dev and test.
Second, for each partition, we convert all relation labels that are assigned to another partition to {\tt NOTA} (steps 3 and 4). Table \ref{tab:transformation_example} shows an example of the transformation process for WIKIDATA.
\begin{table}[t]

    \begin{tabular}{p{7.2 cm}}
        \hline
    \textbf{\textit{Sentence 1:}} ``Among the current participants, Iceland, Norway, and Switzerland are not members of the European Union.''\\

   \textbf{Entity pair:} ``Norway'', ``Switzerland''\\
   \textbf{Original label:} {\tt no\_relation}\\
   \textbf{Label after transformation:} {\tt NOTA}\\
   \textbf{Reason:} {\tt no\_relation} in the supervised setting becomes {\tt NOTA} for FSRE\\
        \hline
    \textbf{\textit{Sentence 2:}} ``Horror writer Stephen King once visited his friend, Peter Straub, whose house is in Crouch End.''\\
    \textbf{Entity pair:} ``Peter Straub'', ``Crouch End''\\
    \textbf{Original label:} {\tt residence}\\
    \textbf{Label after transformation:} {\tt residence}\\
    \textbf{Reason:} The sentence is in the dev set, and the relation label {\tt residence} is part of $\boldsymbol{C}_{dev}$\\
        \hline
    \textbf{\textit{Sentence 3:}} ``Progeny is an American science fiction film released in 1999.''\\
    \textbf{Entity pair:} ``Progeny'', ``American''\\
    \textbf{Original label:} {\tt country of origin}\\
    \textbf{Label after transformation:} {\tt NOTA}\\
    \textbf{Reason:} The sentence is taken from the dev set, but the relation {\tt residence} is part of $\boldsymbol{C}_{test}$\\
        \hline
    \end{tabular}
    \caption{Example data points before and after the transformation process in Algorithm~\ref{alg:transformation}.}
    \label{tab:transformation_example}
\end{table}


\subsection{Episode Sampling}
\label{sec:episodes}

The small number of examples per class in FSRE ($K$) may introduce statistical instability in the results observed. To address this,
episodic learning repeats the training/evaluation of a given method over a large number of episodes that sample different support sentences for the given classes. 
More formally, for a $N$-way $K$-shot setup an episode \(E\) consists of three items:
\begin{enumerate}[label=(\alph*)]
    \item \(N\) randomly chosen target relations:
    \[\boldsymbol{C}_{target} = \{c_1, c_2,....., c_N\} \text{ s.t. } c_{1..N}\notin \{{\tt NOTA}\}\]
    
    \item A randomly chosen support set of size \(K\) for each of the \(N\) relations:
    \[\boldsymbol{X}_{supt} = \{\boldsymbol{X}_1, \boldsymbol{X}_2, .....,\boldsymbol{X}_i,....., \boldsymbol{X_N}\}\]
    \[\boldsymbol{X}_i = \{(x_1, c_i), (x_2, c_i), ...,(x_j, c_i),.., (x_K, c_i)\}\]
    \item A randomly chosen labeled example as a query \(Q = (x_q,c_q)\) such that \((x_q,c_q) \notin \boldsymbol{X}_{supt}\). 
\end{enumerate}

Given an episode \(E = (\boldsymbol{C}_{target}, \boldsymbol{X}_{supt}, Q)\), the goal of a Few-Shot learning classifier is to create a decision function to choose a label from \({C}_{target} \cup \{{\tt NOTA}\}\) for the given query \(Q\). 

We describe a general approach of $N$-Way $K$-Shot episode sampling procedure in the Algorithm \ref{alg:episodeSampling}, where $\boldsymbol{D}_{E}$ and $\boldsymbol{C}_{E}$ are input dataset and labels to sample from, and $\boldsymbol{E}_{test}$ is the set of returned episodes. 

A similar episode sampling approach has been described in \cite{sabo2021revisiting}. To create \textit{train} episodes, \(D_{train}\) and \(C_{train}\) should be used as input. In the same way, \textit{dev} and \textit{test} episodes can be created using their respective data and relation sets. 


\begin{algorithm}[t]
\caption{Episode sampling for a $N$-way $K$-shot FSRE}\label{alg:episodeSampling}
\begin{algorithmic}
\State \textbf{Input: }\(\boldsymbol{D}_{E}, \boldsymbol{C}_{E}, episodeSize\)
\State \textbf{Output: }\(\boldsymbol{E}_{test}\)
\State $\boldsymbol{E}_{test} = \{\}$
\State \(\boldsymbol{C}_{E}' = \boldsymbol{C}_{E} - \{{\tt NOTA}\}\)
\For{\(e=0 \text{ to } episodeSize\)}
    \State \( \boldsymbol{C}_{target} = RandomSample(\boldsymbol{C}_{E}', N)\)
    \State \(\boldsymbol{X}_{supt} = [~]\)
    \For{\(i=0 \text{ to } |{C}_{target}|\)}
    \State \(r = {C}_{target}[i]\)
    \State \( \boldsymbol{X}_i = RandomSample(\boldsymbol{D}_{E}, K, r)\}\)
    \State \(\boldsymbol{X}_{supt}[i] = \boldsymbol{X}_i\)
    \EndFor
    \State \(\boldsymbol{D}_E' = \{\boldsymbol{D}_E: \boldsymbol{D}_E \notin \boldsymbol{X}_{supt}\}\)
    \State \( Q = RandomSample(\boldsymbol{D}_{E}', 1)\)
    \State \(\boldsymbol{C}_{target}' = \boldsymbol{C}_{target} \cup \{{\tt NOTA}\}\) 
    \State $\boldsymbol{E}_{test} = \boldsymbol{E}_{test} \cup \{(\boldsymbol{C}_{target}', {X}_{supt}, Q)\}$
\EndFor\\

\Return \(\boldsymbol{E}_{test}\)
\end{algorithmic}
\end{algorithm}

\section{Dataset Statistics}
\label{sec:dataset-statistics}
Table~\ref{tab:statistics} summarizes key statistics for the three supervised datasets that serve as the starting point for FSRE. We chose three datasets with a significant variation in the number of relations. Table~\ref{tab:statistics} shows that TACRED has 42 relations, NYT29 has 29 relations, and WIKIDATA has  352 relations, which is much larger. Additionally, when we look at the \texttt{NOTA} instances, these three datasets differ enormously. For instance, in the supervised NYT29, there are no \texttt{NOTA} instances. In supervised TACRED, the number of \texttt{NOTA} instances is higher than the number of \texttt{NOTA} instances in the supervised WIKIDATA.  
Table ~\ref{tab:FS_statistics} summarizes how the number of relation instances and \texttt{NOTA} instances in three resulting $FS$ datasets have been changed from supervised datasets. In Appendix B, we present further statistics and analysis demonstrating that our FSRE meta-dataset meets all the requirements of a realistic few-shot relation extraction dataset.

\begin{table}[H]
\begin{small}
\resizebox{.48\textwidth}{!}{  
    \begin{tabular}{lrrr}
        \hline
        &TACRED & NYT29 & WIKIDATA\\
        \hline
        Train Size & 68,124 & 78,885 & 775,919\\

        Dev Size & 22,631 & 5859 &  251,802\\

        Test Size & 15,509 & 8759 & 739,408\\
  
        Relation Class & 42 & 29 & 352\\
  
 
        Relation Instances& 21,773& 93,503 & 1,299,085\\
      
        {\tt NOTA} Instances &84,491 & 0 & 468,044\\
        \hline
    \end{tabular}
    } 
\end{small}
    \caption{Statistics of the supervised TACRED, NYT29, and WIKIDATA datasets. The first three rows report the number of sentences per partition.}
    \label{tab:statistics}
\end{table}

\begin{table}[H]
\begin{small}
\resizebox{.48\textwidth}{!}{  
    \begin{tabular}{lrrr}
        \hline
        &TACRED & NYT29 & WIKIDATA\\
        \hline

        Relation Instances& 9,600 & 58,841 & 513,891\\
      
        NOTA Instances &96,664 & 34,662 & 1,253,238\\
        \hline
    \end{tabular}
    } 
\end{small}
    \caption{Statistics of the Few-Shot TACRED, NYT29, and WIKIDATA datasets.}
    \label{tab:FS_statistics}
\end{table}

\section{Experimental Results}
\label{sec:experiment}
\begin{table*}
    \resizebox{\textwidth}{!}{    
\begin{tabular}{@{\extracolsep{4pt}}lrrrrrr}
\toprule   
{Model} & \multicolumn{3}{c}{5-way 1-shot}  & \multicolumn{3}{c}{5-way 5-shot}\\
 \cmidrule{2-4} 
 \cmidrule{5-7} 
& \multicolumn{1}{c}{P} & \multicolumn{1}{c}{R} & \multicolumn{1}{c}{F1} & \multicolumn{1}{c}{P} & \multicolumn{1}{c}{R} & \multicolumn{1}{c}{F1} \\ 
\midrule
Unsupervised Baseline & 5.70 $\pm$ 0.10 & 91.02 $\pm$ 0.65 & 10.73 $\pm$ 0.18     & 5.65 $\pm$ 0.11 & 95.56 $\pm$ 0.70 & 10.67 $\pm$ 0.20 \\
\midrule
Sentence-Pair \footnotesize{(not fine-tuned)} & 3.9 $\pm$ 0.21 & 5.21 $\pm$ 0.31 & 4.45 $\pm$ 0.24 & 2.76 $\pm$ 0.16 & 8.79 $\pm$ 0.58 & 4.2 $\pm$ 0.25 \\
Sentence-Pair \footnotesize{(fine-tuned)} & 6.89 $\pm$ 0.33 & 28.56 $\pm$ 1.67 & 11.10 $\pm$ 0.55 & 14.94 $\pm$ 0.26 & 24.03 $\pm$ 0.32 & 18.42 $\pm$ 0.16 \\
MNAV          & 15.11 $\pm$ 0.46 & 8.47 $\pm$ 0.31 & 10.85 $\pm$ 0.29 & 24.48 $\pm$ 1.02 & 32.00 $\pm$ 1.07  & 27.73 $\pm$ 0.94 \\
\midrule
OdinSynth     & 23.48 $\pm$ 1.46 & 11.46 $\pm$ 1.02 & 15.40 $\pm$ 1.21 & 29.77 $\pm$ 0.83 & 20.34 $\pm$ 0.53 & 24.16 $\pm$ 0.44 \\

\midrule
Hard-matching Rules  & 51.35 $\pm$ 6.53 & 2.94 $\pm$ 0.48 & 5.56 $\pm$ 0.90           & 45.94 $\pm$ 5.31 & 10.81 $\pm$ 1.23 & 17.50 $\pm$ 1.98 \\
Soft-matching Rules               & 33.46 $\pm$ 1.47 & 19.69 $\pm$ 1.14 & \textbf{24.78 $\pm$ 1.22} & 51.66 $\pm$ 1.85 & 26.02 $\pm$ 1.29 & \textbf{34.59 $\pm$ 1.24} \\
\bottomrule
\end{tabular}

}
  \caption{The results for the 5-way 1-shot and 5-way 5-shot settings on the test partition of the FS TACRED dataset.}
\label{tab:tacred}
  \end{table*}

\begin{table*}
    \resizebox{\textwidth}{!}{    
\begin{tabular}{@{\extracolsep{4pt}}lrrrrrr}
\toprule   
{Model} & \multicolumn{3}{c}{5-way 1-shot}  & \multicolumn{3}{c}{5-way 5-shot}\\
 \cmidrule{2-4} 
 \cmidrule{5-7} 
& \multicolumn{1}{c}{P} & \multicolumn{1}{c}{R} & \multicolumn{1}{c}{F1} & \multicolumn{1}{c}{P} & \multicolumn{1}{c}{R} & \multicolumn{1}{c}{F1} \\ 
\midrule
Unsupervised Baseline & 11.60 $\pm$ 0.18 &40.34 $\pm$ 0.54 & 18.03 $\pm$ 0.26     & 11.70 $\pm$ 0.25 & 40.65 $\pm$ 0.45 & 18.17 $\pm$ 0.34 \\
\midrule
Sentence-Pair \footnotesize{(not fine-tuned)} & 10.61 $\pm$ 0.32 & 12.39 $\pm$ 0.41 & 11.43 $\pm$ 0.35 & 15.81 $\pm$ 0.94 & 5.41 $\pm$ 0.25 & 8.06 $\pm$ 0.39 \\
Sentence-Pair \footnotesize{(fine-tuned)} & 38.09 $\pm$ 2.42 & 7.4 $\pm$ 0.42 & 12.4 $\pm$ 0.71 & 36.48 $\pm$ 1.37 & 16.02 $\pm$ 0.41 & \textbf{22.26 $\pm$ 0.62} \\
MNAV          & 25.08 $\pm$ 0.73& 34.37  $\pm$ 0.87 & \textbf{29.00 $\pm$ 0.80} & 33.24 $\pm$ 1.06& 15.47 $\pm$ 0.38  &  21.12 $\pm$ 0.55 \\
\midrule
OdinSynth     & 30.07 $\pm$ 0.93 & 9.42 $\pm$ 0.31 & 14.34 $\pm$ 0.46 & 21.61 $\pm$ 0.61 & 17.98 $\pm$ 0.45 & 19.63 $\pm$ 0.51 \\
\midrule
Hard-matching Rules  & 77.47 $\pm$ 1.53 & 1.53 $\pm$ 0.13 & 3.01 $\pm$ 0.25           & 80.49 $\pm$ 1.73 & 3.40 $\pm$ 0.12 & 6.52 $\pm$ 0.23 \\
Soft-matching Rules              & 20.80 $\pm$ 0.38 & 12.27 $\pm$ 0.39 & 15.44 $\pm$ 0.40 & 24.50 $\pm$ 0.83 & 16.67 $\pm$ 0.49 & 19.84 $\pm$ 0.59 \\
\bottomrule
\end{tabular}

}
  \caption{The results for the 5-way 1-shot and 5-way 5-shot settings on the test partition of the FS NYT dataset.}
\label{tab:nyt}
  \end{table*}
\begin{table*}[ht!]
    \resizebox{\textwidth}{!}{    
\begin{tabular}{@{\extracolsep{4pt}}lrrrrrr}
\toprule   
{Model} & \multicolumn{3}{c}{5-way 1-shot}  & \multicolumn{3}{c}{5-way 5-shot}\\
 \cmidrule{2-4} 
 \cmidrule{5-7} 
& \multicolumn{1}{c}{P} & \multicolumn{1}{c}{R} & \multicolumn{1}{c}{F1} & \multicolumn{1}{c}{P} & \multicolumn{1}{c}{R} & \multicolumn{1}{c}{F1} \\ 
\midrule
Unsupervised Baseline & 2.52 $\pm$ 0.16  & 29.99 $\pm$ 1.42 & 4.64 $\pm$ 0.28      & 2.28 $\pm$ 0.13  & 54.35 $\pm$ 1.03  & 4.38 $\pm$ 0.24  \\
\midrule
Sentence-Pair \footnotesize{(not fine-tuned)} & 6.4 $\pm$ 1.51 & 2.55 $\pm$ 0.66 & 3.65 $\pm$ 0.92 & 2.68 $\pm$ 0.56 & 8.67 $\pm$ 1.57 & 4.09 $\pm$ 0.82 \\
Sentence-Pair \footnotesize{(fine-tuned)} & 6.65 $\pm$ 0.78 & 7.99 $\pm$ 0.93 & 7.26 $\pm$ 0.85 & 5.76 $\pm$ 0.87 & 8.74 $\pm$ 0.95 & 6.94 $\pm$ 0.93 \\
MNAV          & 17.49 $\pm$ 1.45& 6.76 $\pm$ 1.21 & \textbf{9.74 $\pm$ 1.47} & 15.27 $\pm$ 0.98 & 28.26 $\pm$ 0.96& \textbf{19.83 $\pm$ 1.06} \\
\midrule
OdinSynth     & 12.99 $\pm$ 1.67 & 6.15 $\pm$ 0.58 & 8.34 $\pm$ 0.85 & 10.09 $\pm$ 1.31 & 19.18 $\pm$ 1.57 & 13.21 $\pm$ 1.46 \\
\midrule
Hard-matching Rules  & 6.38 $\pm$ 3.24 & 0.38 $\pm$ 0.20 & 0.72 $\pm$ 0.37           & 5.15 $\pm$ 1.83 & 1.13 $\pm$ 0.37 & 1.85 $\pm$ 0.61 \\
Soft-matching Rules               & 35.88 $\pm$ 10.01 & 2.73 $\pm$ 0.86 & 5.06 $\pm$ 1.58 & 17.58 $\pm$ 3.28 & 9.71 $\pm$ 2.15 & 12.50 $\pm$ 2.59 \\
\bottomrule
\end{tabular}

}
  \caption{The results for the 5-way 1-shot and 5-way 5-shot settings on the test partition of the FS WIKIDATA dataset.}
\label{tab:wikidata}
  \end{table*}

\subsection{Experimental Setup}
We applied the transformation techniques outlined in \S~\ref{sec:supervised-to-fewshot} and \S~\ref{sec:episodes} on all datasets described in the previous section to produce their FSRE variants.\footnote[3]{To enable comparison with previous work, for TACRED we kept the transformation introduced in~\cite{sabo2021revisiting}.}
We tested on all datasets in $5$-way $1$-shot and $5$-way $5$-shot scenarios. In both cases, we repeat the procedure with 5 different random seeds.

\subsection{Models}
\label{sec:models}

We evaluated the following baselines and models:

\paragraph{Unsupervised Baseline} -- 
This baseline model uses {\em solely} the entity types in both the query sentence and the support sentences for classification during inference \cite{vacareanu-etal-2022-examples}. If there are support sentences with the same entity types as the query sentence, the model randomly chooses one and predicts its relation. In other cases, the baseline predicts \texttt{NOTA}.

\paragraph{Sentence-Pair} --
We implement a baseline similar to \citelanguageresource{Gao2019FewRel2T}, which operates as follows: 
We pair each query sentence to each support sentence and feed the concatenated text to a sentence transformer \cite{Reimers2019SentenceBERTSE} to obtain a single score that quantifies the degree to which both sentences convey the same underlying relation. 
During training, we fine-tune the model to maximize the score between sentences with the same relation and minimize the score between sentences with different relation (or \texttt{NOTA}). 
During inference, we predict the relation associated with the highest score, provided it is above a threshold tuned on the development partition. Otherwise, we predict \texttt{NOTA}.
We use a pre-trained model and show results with and without fine-tuning.\footnote[4]{\texttt{cross-encoder/ms-marco-MiniLM-L-6-v2}}


\paragraph{MNAV} --
Multiple {\tt NOTA} Vectors (MNAV) is an extended version of the NAV method, which computes a score between the query vector, each support sentence vector, and, additionally, a learned vector for the NOTA class \cite{sabo2021revisiting}. Instead of just one vector for NOTA, MNAV uses multiple vectors to account for the fact that NOTA is a ``catch all'' for all other relations. The number of NOTA vectors is tuned on the development set. In the classification process, the model selects the nearest vector to the query representation to establish the predicted relation label. 

\paragraph{OdinSynth} -- 
 OdinSynth is a transformer-based rule synthesis model that generates rules from the provided support sentences and then applies these rules to the query sentence \cite{vacareanu-etal-2022-examples}. If none of the rules match, the model predicts \texttt{NOTA}. If there exists a match with one or more rules, the model predicts the relation through majority voting.
 
\paragraph{Hard-Matching Rules} --
Represent lexico-syntactic rules created over the shortest path connecting the two entities.

\paragraph{Soft-Matching Rules} --
This is a neuro-symbolic model \cite{Vacareanu2024BestOB} that aims to increase the recall of rules by leveraging the high expressivity of neural networks. The method first attempts to match a rule the traditional way (see \textit{Hard-Matching Rules}). If the match fails, it then falls back to the neural component, which will predict a matching score $s \in [0, 1]$. The training of the neural component utilizes (rule, sentence) tuples along with a contrastive loss function. The objective is to maximize the similarity between rules and sentences with the same relation while minimizing it for those with different relations. 

In addition to a comprehensive assessment of the six recent few-shot relation extraction models mentioned above, we also evaluated a Zero-Shot Large Language Model (LLM) baseline on our FSRE meta-dataset. The details of this baseline and the experimental result are provided in Appendix C.

\subsection{Results Analysis}
\label{sec:ResultAnalysis}
Table \ref{tab:tacred}, \ref{tab:nyt}, \ref{tab:wikidata} represent the result of different models on our resulting FSRE datasets. 
We draw the following conclusions:

First, no single method emerges as the clear top performer across all scenarios. For instance, \textit{Soft-matching Rules} achieves the highest performance on Few-Shot TACRED (Table~\ref{tab:tacred}), \textit{MNAV} excels on Few-Shot WIKIDATA (Table~\ref{tab:wikidata}), and in the case of Few-Shot NYT, \textit{MNAV} performs best for 1-shot, while \textit{Sentence-Pair} leads for 5-shot (Table~\ref{tab:nyt}). The latter result is surprising, given the simplicity of this baseline.

Second, the performance varies drastically between the datasets for both 1-shot and 5-shot scenarios. For instance, in Few-Shot WIKIDATA 5-way 1-shot, the top-performing method achieves an F1 score of $9.74$, whereas in Few-Shot TACRED 5-way 1-shot, the best method reaches an F1 score of $24.78$. This underscores the importance of employing multiple evaluation datasets to gain a realistic assessment of a model's performance. Further, the overall performance across all datasets is low, which indicates a substantial need for future research in this domain. 

In our evaluation of the six models, FS WIKIDATA exhibited comparatively lower performance across all datasets. To understand the underlying reasons, we conducted a qualitative error analysis on FS WIKIDATA, the details of which are provided in Appendix D.



\label{sec:analysis}

\section{Conclusion} 
In this paper, we presented a meta dataset for few-shot relation extraction (FSRE), which comprises three FSRE datasets: two were derived from established supervised relation extraction datasets, while one is an existing FSRE dataset.
All datasets were intricately derived to replicate real-world scenarios, ensuring a strong alignment with real-world contexts. Then, we assessed six relation extraction methods on this meta dataset and found that no single model consistently performs well across all scenarios. This suggests the need for future research in this domain.

As future work, we plan to leverage the resulting dataset to develop methods that demonstrate consistent and robust performance.


\section{Acknowledgements}
This work was partially supported by the Defense Advanced Research Projects Agency (DARPA) under the Habitus and Automating Scientific Knowledge Extraction and Modeling (ASKEM) programs. Mihai Surdeanu declares a financial interest in \url{lum.ai}. This interest has been properly disclosed to the University of Arizona Institutional Review Committee and is managed in accordance with its conflict of interest policies.

\nocite{*}
\section{Bibliographical References}\label{sec:reference}

\bibliographystyle{lrec-coling2024-natbib}
\bibliography{lrec-coling2024}

\section{Language Resource References}
\label{lr:ref}
\bibliographystylelanguageresource{lrec-coling2024-natbib}
\bibliographylanguageresource{languageresource}

\section*{Appendix A}

Table~\ref{tab:supervised_dataset_details} summarizes the tabular format used to represent the three supervised datasets that are the starting point of the few-shot datasets generated in this work.

\begin{table}[H]
    \begin{tabular}{|p{2.3cm}|p{5cm}|}
        \hline
        Field & Description \\
        \hline
        id & Incremental unique ID for each example
        or sentence \\
        \hline
        docid & For dev set docid = "dev", for test set docid= "test", and for train set docid = "train"\\
        \hline
        relation & This field denotes the relation labels between the given entities \\
        \hline
        token & An instance of a sequence or word in the sentence  \\
        \hline
        subj\char`_start & Start index of the subject in a sentence \\
        \hline
        subj\char`_end & End index of the subject in a sentence  \\
        \hline
        obj\char`_start & Start index of the object in a sentence  \\
        \hline
        obj\char`_end & End index of the object in a sentence  \\
        \hline
        subj\char`_type & Subject type (e.g., person name) in a sentence  \\
        \hline
        obj\char`_type & Object type (e.g., person name) in a sentence   \\
        \hline
        stanford\char`_pos & POS tag of the current token   \\
        \hline
        stanford\char`_ner & Named entity label of the current token   \\
        \hline
        stanford\char`_head & 1-based index of the dependency head of the current token   \\
        \hline
        stanford\_deprel & dependency relation of the current token to its head token  \\
        \hline
        
    \end{tabular}
    \caption{Descriptions of the columns in the tabular format used to encode the three supervised datasets used in this work. Note that the ``stanford'' prefix for the last three columns is maintained for compatibility with the TACRED format; in NYT29 and WIKIDATA, this information is generated using the {\tt processors} library instead.}
    \label{tab:supervised_dataset_details}
\end{table}

\section*{Appendix B}
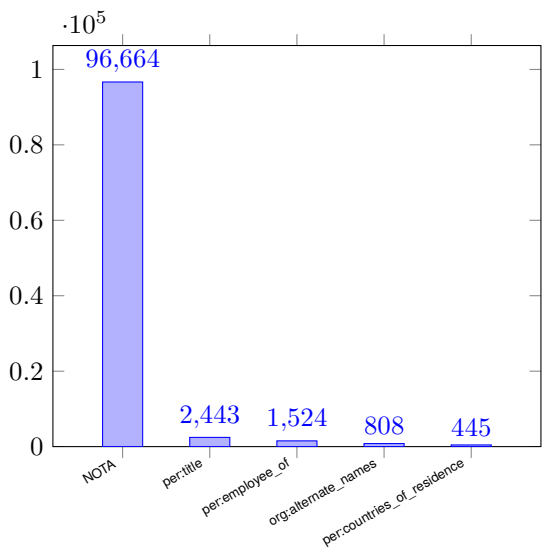
\begin{figure}
\centering
\begin{tikzpicture}
\begin{axis}[
    ybar,
    bar width=15pt,
    xlabel style={at={(axis description cs:1.05,-0.1)}},
    ylabel style={at={(axis description cs:-0.15,1.05)}},
    symbolic x coords={NOTA, per:title, per:employee\_of, org:alternate\_names, per:countries\_of\_residence},
    xtick=data,
    x tick label style={font=\tiny, rotate=30, anchor=east},
    nodes near coords,
    nodes near coords align={vertical},
    ymin=0,
    enlarge x limits=0.2
]
\addplot coordinates {
    (NOTA, 96664)
    (per:title, 2443)
    (per:employee\_of, 1524)
    (org:alternate\_names, 808)
    (per:countries\_of\_residence, 445)
};
\end{axis}
\end{tikzpicture}
\caption{Few-Shot TACRED top five relation distribution} \label{fig:tacred-rel-Dist}
\end{figure}

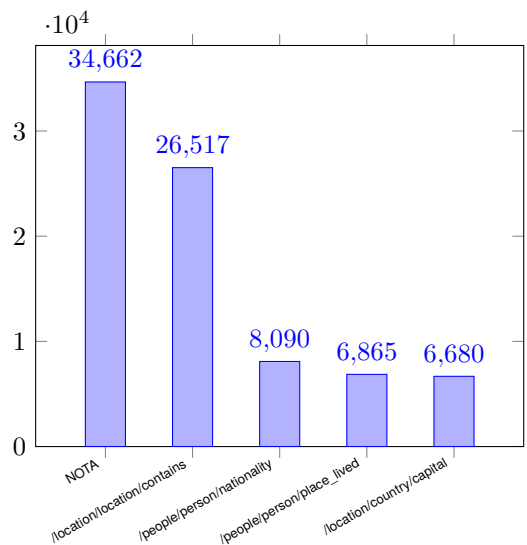
\begin{figure}
\centering
\begin{tikzpicture}
\begin{axis}[
    ybar,
    bar width=15pt,
    xlabel style={at={(axis description cs:1.05,-0.1)}},
    ylabel style={at={(axis description cs:-0.15,1.05)}},
    symbolic x coords={NOTA, /location/location/contains, /people/person/nationality, /people/person/place\_lived, /location/country/capital},
    xtick=data,
    x tick label style={font=\tiny, rotate=30, anchor=east},
    nodes near coords,
    nodes near coords align={vertical},
    ymin=0,
    enlarge x limits=0.2
]
\addplot coordinates {
    (NOTA, 34662)
    (/location/location/contains, 26517)
    (/people/person/nationality, 8090)
    (/people/person/place\_lived, 6865)
    (/location/country/capital, 6680)
};
\end{axis}
\end{tikzpicture}
\caption{Few-Shot NYT29 top five relation distribution} \label{fig:nyt29-rel-Dist}
\end{figure}

\begin{figure}
\centering
\begin{tikzpicture}
\begin{axis}[
    ybar,
    bar width=15pt,
    xlabel style={at={(axis description cs:1.05,-0.1)}},
    ylabel style={at={(axis description cs:-0.15,1.05)}},
    symbolic x coords={NOTA, country, located in the administrative territorial entity, instance of, shares border with},
    xtick=data,
    x tick label style={font=\tiny, rotate=30, anchor=east},
    nodes near coords,
    nodes near coords align={vertical},
    ymin=0,
    enlarge x limits=0.2
]
\addplot coordinates {
    (NOTA, 1253238)
    (country, 81394)
    (located in the administrative territorial entity, 63773)
    (instance of, 38796)
    (shares border with, 37290)
};
\end{axis}
\end{tikzpicture}
\caption{Few-Shot WIKIDATA top five relation distribution} \label{fig:wikidata-rel-Dist}
\end{figure}
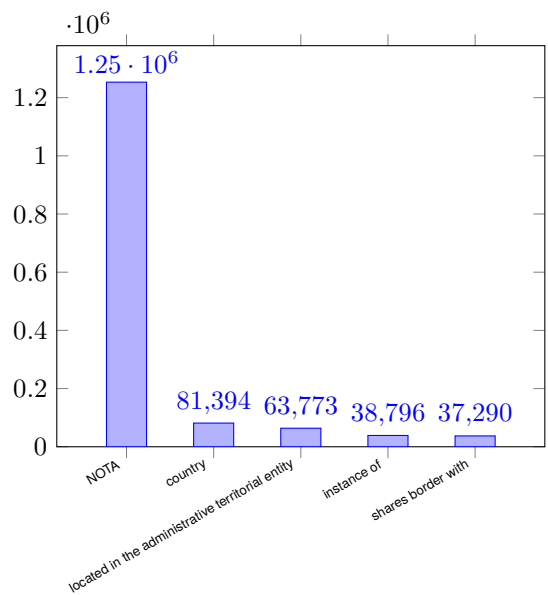

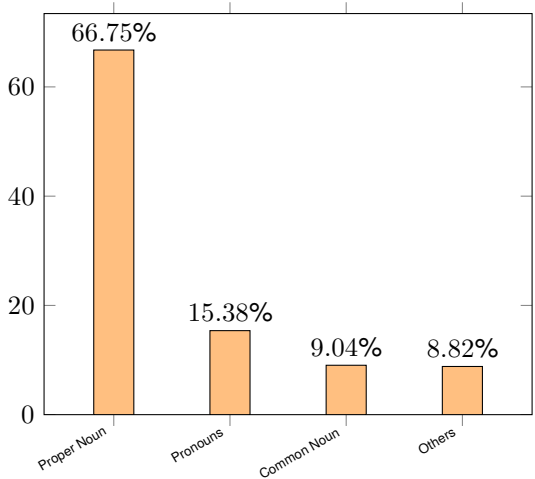
\begin{figure}
\centering
\begin{tikzpicture}
\begin{axis}[
    ybar,
    bar width=15pt,
    xlabel style={at={(axis description cs:1.05,-0.1)}},
    ylabel style={at={(axis description cs:-0.15,1.05)}},
    symbolic x coords={Proper Noun, Pronouns, Common Noun, Others},
    xtick=data,
    x tick label style={font=\tiny, rotate=30, anchor=east},
    nodes near coords={\pgfmathprintnumber\pgfplotspointmeta\%} ,
    nodes near coords align={vertical},
    ymin=0,
    enlarge x limits=0.2
]
\addplot[fill = orange!50] coordinates {
    (Proper Noun, 66.75)
    (Pronouns, 15.38)
    (Common Noun, 9.043)
    (Others, 8.82)
};
\end{axis}
\end{tikzpicture}
\caption{Few-Shot TACRED Entity POS tag distributions} \label{fig:tacred-pos-dist}
\end{figure}

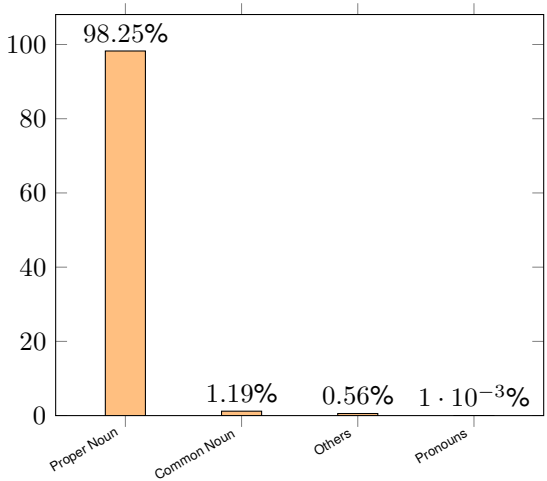
\begin{figure}
\centering
\begin{tikzpicture}
\begin{axis}[
    ybar,
    bar width=15pt,
    xlabel style={at={(axis description cs:1.05,-0.1)}},
    ylabel style={at={(axis description cs:-0.15,1.05)}},
    symbolic x coords={Proper Noun, Common Noun, Others, Pronouns},
    xtick=data,
    x tick label style={font=\tiny, rotate=30, anchor=east},
    nodes near coords={\pgfmathprintnumber\pgfplotspointmeta\%} ,
    nodes near coords align={vertical},
    ymin=0,
    enlarge x limits=0.2
]
\addplot[fill = orange!50] coordinates {
    (Proper Noun, 98.25)
    (Common Noun, 1.19)
    (Others, 0.56)
    (Pronouns, 0.001)
};
\end{axis}
\end{tikzpicture}
\caption{Few-Shot NYT29 Entity POS tag distributions} \label{fig:nyt-pos-dist}
\end{figure}

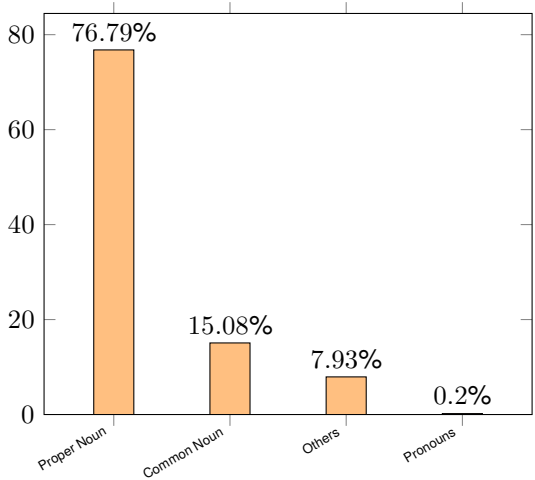
\begin{figure}
\centering
\begin{tikzpicture}
\begin{axis}[
    ybar,
    bar width=15pt,
    xlabel style={at={(axis description cs:1.05,-0.1)}},
    ylabel style={at={(axis description cs:-0.15,1.05)}},
    symbolic x coords={Proper Noun, Common Noun, Others, Pronouns},
    xtick=data,
    x tick label style={font=\tiny, rotate=30, anchor=east},
    nodes near coords={\pgfmathprintnumber\pgfplotspointmeta\%} ,
    nodes near coords align={vertical},
    ymin=0,
    enlarge x limits=0.2
]
\addplot[fill = orange!50] coordinates {
    (Proper Noun, 76.79)
    (Common Noun, 15.08)
    (Others, 7.93)
    (Pronouns, 0.20)
};
\end{axis}
\end{tikzpicture}
\caption{Few-Shot WIKIDATA Entity POS tag distributions} \label{fig:wikidata-pos-dist}
\end{figure}

In section \ref{sec:supervised-to-fewshot}, we outlined six characteristics essential for a realistic Few-Shot dataset. Our FSRE meta dataset fulfills all these six criteria. 
For instance, we split the original dataset into three partitions (train/dev/test) such that they are pairwise disjoint with respect to the positive relations they contain (as outlined in Steps 1 and 2 of Algorithm 1). This guarantees that the test relation classes in our dataset are distinct from any relations that might be present in a background dataset, thereby fulfilling constraint (a) of the realistic Few-Shot assumption. Moreover, we follow the $5$-way $1$-shot and $5$-way $5$-shot setup for episode sampling, which ensures that the number of training examples for each target relation class is very small (1 or 5), thus satisfying constraint (b). Figure \ref{fig:tacred-rel-Dist}, \ref{fig:nyt29-rel-Dist}, \ref{fig:wikidata-rel-Dist} illustrate the non-uniformity of the relation classes and the predominance of {\tt NOTA} class, indicative of satisfying realistic constraints (c), (d), and (e). Figure \ref{fig:tacred-pos-dist}, \ref{fig:nyt-pos-dist}, \ref{fig:wikidata-pos-dist} indicate the presence of a variety of POS tags, with a notable percentage of proper nouns, common nouns, and pronouns, reflecting the diversity and realism of entity distributions, thus satisfying constraint (f).

\section*{Appendix C}
\subsection*{Zero-Shot LLM Baseline}
We evaluated the Zero-Shot relation classification performance of the Large Language Model (LLM) using \textit{GPT-4}. The experiment was conducted on a test set containing ten episodes, with each episode containing three test sentences. For each sentence, we prompted \textit{GPT-4} to identify a relation for a given entity pair using the prompting technique described by \citet{Zhang2023LLM-QA4RE}. The prompt includes the label verbalization technique to articulate the relations. We conducted the experiment in both $5$-way $1$-shot and $5$-way $5$-shot configurations, where \textit{GPT-4} was tasked with classifying the relation for the given entity pair into one of the five target relations or indicating `None of the Above' ({\tt NOTA}) if none is applicable. Figure \ref{fig:zero-shot-llm-prompt} illustrates an example of the prompt. 

The results of the experiment are presented in Tables \ref{tab:tacredZeroShotLLM}, \ref{tab:nytZeroShotLLM} and \ref{tab:wikidataZeroShotLLM}. In the tables, we included the performance scores of other models on the same test set to facilitate easier comparison. The results show that zero-shot LLM achieves low precision and high recall in FS TACRED (see Table \ref{tab:tacredZeroShotLLM}) and FS NYT29 (see Table \ref{tab:nytZeroShotLLM}). The low precision is attributed to a high false positive rate, where \textit{GPT-4} often chose a positive relation from the target set instead of selecting {\tt NOTA} when the correct relation was not among the target relations. However, when the correct relation is included in the target set, \textit{GPT-4} tends to identify it correctly, resulting in a high true positive rate. Although GPT-4 generally performs better than the other models, it is not always the best in every scenario. For example, in the FS NYT29 $5$-way $1$-shot configuration, the \textit{MNAV} model outperforms \textit{GPT-4}, and in the FS WIKIDATA $5$-way $5$-shot setup, the \textit{Unsupervised Baseline} model performs better than GPT-4. This reinforces the conclusion drawn in section \ref{sec:ResultAnalysis} that no single model consistently stands out as the best performer across all scenarios, underscoring the significant need for continued research in this field.

Since a small test set was utilized in this experiment, further research is necessary to gain a deeper understanding of the Zero-Shot relation classification capabilities of Large Language Models.

\begin{table*}[h!]
\resizebox{\columnwidth}{!}{    
\begin{tabular}{@{\extracolsep{4pt}}lrrrrrr}
\toprule   
{Model} & \multicolumn{3}{c}{5-way 1-shot}  & \multicolumn{3}{c}{5-way 5-shot}\\
 \cmidrule{2-4} 
 \cmidrule{5-7} 
& \multicolumn{1}{c}{P} & \multicolumn{1}{c}{R} & \multicolumn{1}{c}{F1} & \multicolumn{1}{c}{P} & \multicolumn{1}{c}{R} & \multicolumn{1}{c}{F1} \\ 
\midrule
Unsupervised Baseline & 8.33 & 33.33 & 13.33 & 11.76 & 66.67 & 20 \\
\midrule
MNAV          & 0 & 0 & 0 & 25 & 33.33  & 28.57 \\

\midrule
Hard-matching Rules  & 0 & 0 & 0 & 0 & 0 & 0 \\
Soft-matching Rules & 66.67 & 66.67 & 66.67 & 16.67 & 33.33 & 22.22 \\
\midrule
Zero-Shot LLM \footnotesize{(GPT 4)} & 50 & 100 & \textbf{67} & 27 & 100 & \textbf{43} \\
\bottomrule
\end{tabular}

}
\caption{The results for the 5-way 1-shot and\\ 5-way 5-shot settings on a small test partition\\ of the FS TACRED dataset.}

\label{tab:tacredZeroShotLLM}
  \end{table*}

\begin{table*}[h!]
\resizebox{\columnwidth}{!}{    
\begin{tabular}{@{\extracolsep{4pt}}lrrrrrr}
\toprule   
{Model} & \multicolumn{3}{c}{5-way 1-shot}  & \multicolumn{3}{c}{5-way 5-shot}\\
 \cmidrule{2-4} 
 \cmidrule{5-7} 
& \multicolumn{1}{c}{P} & \multicolumn{1}{c}{R} & \multicolumn{1}{c}{F1} & \multicolumn{1}{c}{P} & \multicolumn{1}{c}{R} & \multicolumn{1}{c}{F1} \\ 
\midrule
Unsupervised Baseline & 18.18 & 50 & 26.67 & 12.5 & 37.50 & 18.75 \\
\midrule
MNAV          & 58.33 & 87.5 & \textbf{69.99} & 10.07 & 55.77  & 17.06 \\

\midrule
Hard-matching Rules  & 0 & 0 & 0 & 0 & 0 & 0 \\
Soft-matching Rules & 25 & 37.5 & 30 & 25 & 37.5 & 30 \\
\midrule
Zero-Shot LLM \footnotesize{(GPT 4)} & 26.08 & 75 & 38.71 & 21.74 & 62.5 & \textbf{32.26} \\
\bottomrule
\end{tabular}

}
\caption{The results for the 5-way 1-shot and\\ 5-way 5-shot settings on a small test partition\\ of the FS NYT29 dataset.}

\label{tab:nytZeroShotLLM}
  \end{table*}


\begin{table*}[h!]
\resizebox{\columnwidth}{!}{    
\begin{tabular}{@{\extracolsep{4pt}}lrrrrrr}
\toprule   
{Model} & \multicolumn{3}{c}{5-way 1-shot}  & \multicolumn{3}{c}{5-way 5-shot}\\
 \cmidrule{2-4} 
 \cmidrule{5-7} 
& \multicolumn{1}{c}{P} & \multicolumn{1}{c}{R} & \multicolumn{1}{c}{F1} & \multicolumn{1}{c}{P} & \multicolumn{1}{c}{R} & \multicolumn{1}{c}{F1} \\ 
\midrule
Unsupervised Baseline & 10 & 10 & 10 & 50 & 58.33 & \textbf{53.85} \\
\midrule
MNAV          & 0 & 0 & 0 & 66.67 & 16.67  & 26.67 \\

\midrule
Hard-matching Rules  & 100 & 0 & 0 & 100 & 0 & 0 \\
Soft-matching Rules & 33.33 & 10 & 15.38 & 33.33 & 10 & 15.38 \\
\midrule
Zero-Shot LLM \footnotesize{(GPT 4)} & 55.55 & 50 & \textbf{52.63} & 60 & 46.15 & 52.17 \\
\bottomrule
\end{tabular}

}
\caption{The results for the 5-way 1-shot and\\ 5-way 5-shot settings on a small test partition\\ of the FS WIKIDATA.}

\label{tab:wikidataZeroShotLLM}
  \end{table*}


\begin{figure}
\centering
\begin{tikzpicture}
    \draw[fill=blue!5, line width=0.5pt, rounded corners=5pt] (0,0) rectangle (\linewidth,10) node[pos=.5] {
    \begin{minipage}{0.9\linewidth}
    \color{black} 
    \fontsize{9pt}{8pt}\selectfont 
    \ttfamily	
Given a sentence and two entities within the sentence, classify the relation between the two entities based on the provided sentence. All possible relations are listed below:\\\\
 -org:top\_members/employees: Entity 1  has the high level member  Entity 2\\\\
 -per:schools\_attended: Entity 1 studied in Entity 2\\\\
 -org:founded\_by: Entity 1 was founded by Entity 2\\\\
 -per:origin: Entity 1 has the nationality Entity 2\\\\
 -per:date\_of\_birth: Entity 1 has birthday on Entity 2\\\\
 -NOTA: None of the above\\\\
 
 Sentence: \enquote{In an atmosphere of conflict and misunderstanding, the travel and tourism industry can be an incredibly powerful force for conciliation,} said PATA president and chief executive officer Peter de Jong.\\\\
 Entity 1: PATA\\
 Entity 2: Peter de Jong\\
     \end{minipage}
    };
\end{tikzpicture}
\caption{An example of prompt for Zero-Shot LLM baseline.} \label{fig:zero-shot-llm-prompt}
\end{figure}
\section*{Appendix D}
\subsection*{Qualitative Error Analysis}

In the few-shot relation extraction (FSRE) setting, the performance of all six models we evaluated was comparably low when evaluated on WIKIDATA. This can be primarily attributed to the high prevalence of long-tail entities in WIKIDATA. In \cite{chen2023knowledge}, it is reported that approximately half of the entities in WIKIDATA fall into the long-tail category. The challenges stemming from this prevalence of long-tail entities contribute significantly to the observed performance degradation. Firstly, the data scarcity inherent in long-tail entities exacerbates the already challenging few-shot learning scenario, where models are expected to generalize from limited examples. With fewer instances available for these long-tail entities, models struggle to capture the diverse range of relation patterns and semantic nuances associated with them. Additionally, the lack of contextual cues and varied semantic contexts surrounding these entities further compounds the difficulty of accurate relation extraction. As a result, the efficacy of models in the FSRE setting is hampered by the combination of data scarcity and the intricate nature of relations involving long-tail entities in WIKIDATA.
\end{document}